\documentclass{article}

\usepackage[preprint]{neurips_2023}
\usepackage[utf8]{inputenc} 
\usepackage[T1]{fontenc}    
\usepackage{url}            
\usepackage{booktabs}       
\usepackage{amsfonts}       
\usepackage{nicefrac}       
\usepackage{microtype}      
\usepackage{xcolor}         
\usepackage{multirow}
\usepackage[hidelinks]{hyperref}
\usepackage{graphicx}
\usepackage{xspace}
\usepackage{geometry}
\usepackage{changepage}
\usepackage{arydshln}
\usepackage{booktabs} 
\usepackage{adjustbox} 
\usepackage{caption}
\usepackage{subcaption}
\usepackage{amssymb}
\usepackage{pifont}

\usepackage{titlesec}

\titleformat{\part}[display]   
  {\normalfont\large\bfseries} 
  {\partname\ \thepart}        
  {1pt}                       
  {}                           

\newcommand{\Mississippi}{\mbox{\textit{H2OVL-Mississippi}}\xspace}
\newcommand{\vlm}{\mbox{\textit{H2OVL-Mississippi-2B}}\xspace}
\newcommand{\vlmtiny}{\mbox{\textit{H2OVL-Mississippi-0.8B}}\xspace}

\title{H2OVL-Mississippi Vision Language Models Technical Report}
\author{
Shaikat Galib\thanks{The first three authors contributed equally.}
$\quad$Shanshan Wang\footnotemark[1]
$\quad$Guanshuo Xu\footnotemark[1]
$\quad$Pascal Pfeiffer \\
$\quad$\textbf{Ryan Chesler} 
$\quad$\textbf{Mark Landry}
$\quad$\textbf{Sri Satish Ambati}\\
H2O.ai \\
\texttt{\{firstname.lastname, sri\}@h2o.ai}
}
\date{}

\begin{document}

\maketitle

\section{Abstract}

Smaller vision-language models (VLMs) are becoming increasingly important for privacy-focused, on-device applications due to their ability to run efficiently on consumer hardware for processing enterprise commercial documents and images. These models require strong language understanding and visual capabilities to enhance human-machine interaction. To address this need, we present \Mississippi, a pair of small VLMs trained on 37 million image-text pairs using 240 hours of compute on 8 × H100 GPUs. \vlmtiny is a tiny model with 0.8 billion parameters that specializes in text recognition, achieving state of the art performance on the Text Recognition portion of OCRBench and surpassing much larger models in this area. Additionally, we are releasing \vlm, a 2 billion parameter model for general use cases, exhibiting highly competitive metrics across various academic benchmarks. Both models build upon our prior work with H2O-Danube language models, extending their capabilities into the visual domain. We release them under the Apache 2.0 license, making VLMs accessible to everyone, democratizing document AI and visual LLMs.

\textbf{\Mississippi model collection:}\newline
\small\url{https://huggingface.co/collections/h2oai/h2ovl-mississippi-66e492da45da0a1b7ea7cf39} 

\section{Introduction}

The field of vision-language models (VLMs) has rapidly evolved, with significant strides made in connecting visual encoders to language models to enhance the capabilities of AI in handling diverse visual and textual tasks. While current state-of-the-art models deliver impressive results, they often depend on large architectures that require extensive computational resources. The \Mississippi models seek to address this limitation by offering efficient, smaller-scale alternatives that can compete with larger models across various vision-language tasks, especially in Optical Character Recognition (OCR) and document analysis. This paper introduces the \vlmtiny and \vlm models, detailing their architecture, training methodology, and performance evaluations to highlight their efficiency and adaptability for real-world multimodal tasks. By adopting a data-driven approach, the \Mississippi models provide a scalable and efficient solution for applications in document understanding and multimodal reasoning.

The development of the \Mississippi models is guided by two primary goals: specialization and versatility. The \vlmtiny model is specifically optimized for OCR and document-centric tasks, to provide high accuracy and efficiency in structured information extraction, even in resource-constrained environments. The \vlm model is designed to be a general-purpose vision-language model, capable of performing a wide range of multimodal tasks such as image captioning, visual question answering (VQA), and reasoning. By combining these two approaches, the \Mississippi series aims to deliver models that are not only task-specific but also versatile enough to adapt to diverse visual and textual challenges, ensuring a comprehensive solution for multimodal AI applications.


\section{Related Works}

Early VLMs focused on connecting vision encoders to language models through trainable connectors, allowing models to align visual and textual representations. Notable examples include Flamingo \cite{alayrac2022flamingo} and BLIP-2 \cite{li2023blip2}, which achieved strong results in tasks such as image captioning and visual question answering (VQA) by leveraging pre-trained vision and language components.

LLaVA \cite{liu2023llava} extended this approach by introducing multimodal instruction tuning, enabling models to follow human instructions across visual tasks, such as interactive dialogue about images. This capability set a new benchmark for multimodal interaction and improved the model’s ability to transfer knowledge across tasks.

Further advancements were made with models like PaLI \cite{chen2023pali}, Florence-2 \cite{yuan2023florence2}, and Unified-IO 2 \cite{lu2023unifiedio}, which jointly trained vision and language components instead of relying on frozen pre-trained encoders. This joint training approach improved the model's performance on complex, cross-modal tasks such as document parsing and visual reasoning.

Decoder-only models, like Fuyu \cite{chen2023fuyu} and CM3 \cite{aghajanyan2023cm3}, streamlined the architecture by using a single transformer to process both image and text inputs. This simplification increased training and inference efficiency, making these models attractive for scenarios where computational resources are limited.

Recently, encoder-decoder models, like Qwen2-VL \cite{qwen2vl} utilize a Naive Dynamic Resolution mechanism, enabling it to process images at varying resolutions by dynamically adjusting the number of visual tokens. This allows the model to handle complex visual tasks such as detailed image captioning and OCR with improved efficiency and accuracy. Similarly, InternVL 1.5 \cite{internvl} adopts a high-resolution strategy, breaking down images into tiles, which improves the model’s ability to capture fine details across a range of vision tasks. Other models, such as Mini-Monkey \cite{huang2023minimonkey}, tackle high-resolution image processing challenges by introducing multi-scale adaptive cropping, which allows models to capture small or irregularly shaped objects more accurately.

The \vlmtiny and \vlm models build on these advancements by utilizing large and diverse datasets to further enhance multimodal performance, ensuring effective handling of a broad range of visual and textual tasks.

\section{Model Architecture}

\begin{figure}[h] 
\centering 
\includegraphics[width=\linewidth]{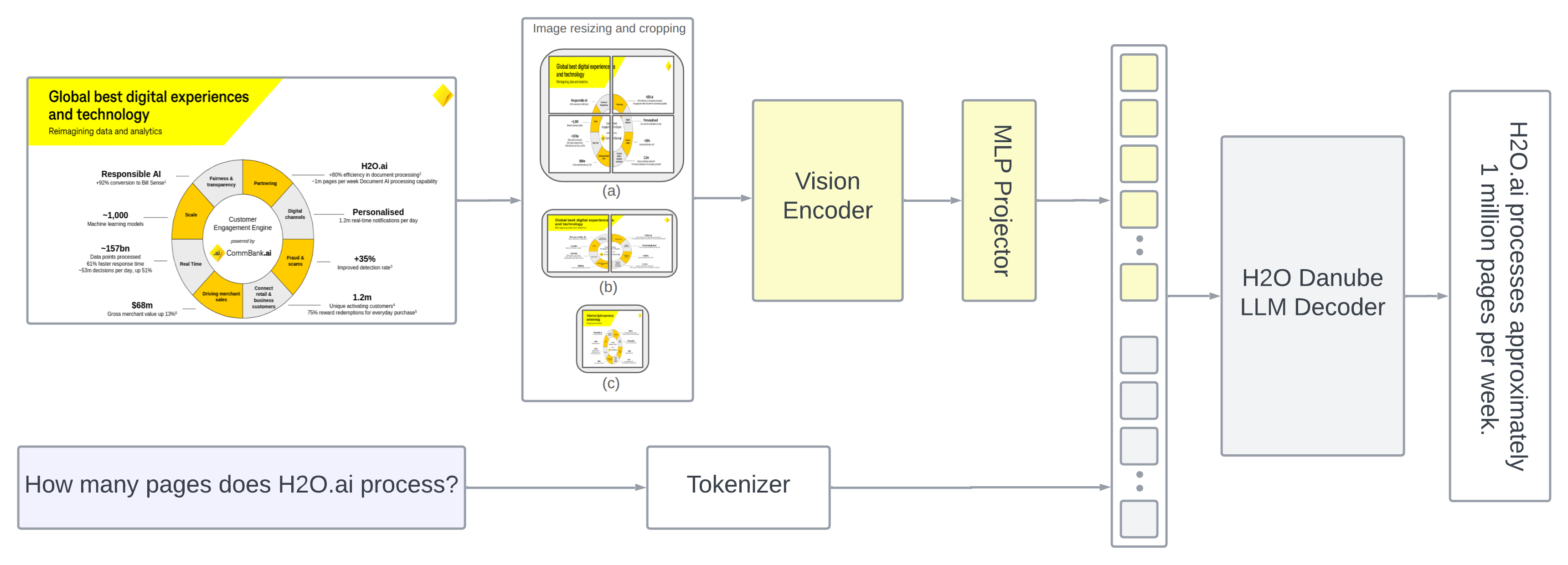} 
\caption{\Mississippi Model Architecture:  The diagram illustrates the procedure for processing input images and text to the LLM. The input image undergoes resizing and cropping at various aspect ratios: (a) Resizing and cropping to the closest original size and aspect ratio, (b) Resizing and cropping to a different aspect ratio, and (c) Resizing the entire image to a fixed 448x448 pixels.}
\label{fig:arch} 
\end{figure}

The architecture of the \Mississippi model takes inspiration from the LLaVA \cite{liu2023llava} and InternVL \cite{internvl} series, following a ViT-MLP-LLM configuration, as shown in Figure~\ref{fig:arch}. It uses a transformer-based setup comprising a vision encoder, an MLP layer, and a large language model (LLM). The vision encoder extracts features from images, while the LLM generates text. The MLP layer acts as a bridge between the vision encoder and the LLM.

Specifically, the \Mississippi architecture integrates the InternViT-300M as its vision encoder and supports two variations for the language model: Danube-2 (1.8 billion parameters) \cite{singer2024h2odanube18btechnicalreport} and Danube-3 (500 million parameters) \cite{pfeiffer2024h2odanube3technicalreport}, providing flexibility based on computational requirements.

The architecture uses a dynamic resolution strategy \cite{internvl} that adjusts image processing based on the image's aspect ratio and resolution. It divides each image into 448x448 pixel tiles, using between 1 and 6 tiles for full coverage of the image (Figure 1a). During training, the number of tiles varies, producing 256 to 1,590 visual tokens, allowing the model to adapt to different image dimensions while optimizing token usage and preserving key details.

To enhance computational efficiency, the architecture incorporates a pixel shuffle operation applied to the Vision Transformer (ViT) embeddings, reducing the number of visual tokens per 448x448 tile to 256. Typically used in image super-resolution tasks to rearrange and combine pixels from low-resolution images, pixel shuffling is adapted here to efficiently decrease the token count while maintaining significant information from each tile. This adaptation ensures effective processing of high-resolution images with reduced computational demands.

Furthermore, the \vlm model uses a multi-scale adaptive cropping (MSAC) strategy, as outlined in the Mini-Monkey report \cite{huang2023minimonkey}. MSAC addresses the sawtooth effect \cite{huang2023minimonkey}, a common issue in traditional cropping techniques, by generating multi-scale representations. This capability enables the model to capture features at different scales, improving performance on tasks involving small or irregularly shaped objects, such as document parsing and image recognition. Similar to the dynamic resolution strategy, MSAC varies the number of tiles from 2 to 6, as illustrated in Figure 1(b).

Finally, a resized version of the original image, scaled to 448x448 pixels, is included in the set of tiles to provide the model with a complete view of the image, improving its ability to capture the overall layout information (Figure 1(c)).

These advanced image processing techniques enable the model to balance efficiency and visual detail, ensuring strong performance across multimodal tasks. The dynamic resolution and MSAC strategies allow it to adapt to diverse image sizes and aspect ratios, optimizing token use while preserving image context. This versatility makes \Mississippi a scalable and effective solution for tasks that require information extraction from fine-grained images.

\section{Training Methodology}

Training a vision language model involves learning complex relationships between images and corresponding texts by jointly optimizing a pre-trained vision encoder (ViT), a pre-trained language model (LLM), and a randomly initialized MLP projector that connects the two. LLaVA \cite{liu2023llava} demonstrated that pre-training the connector with image-caption pairs significantly enhances performance outcomes. Qwen2-VL \cite{qwen2vl} highlighted the benefits of pre-training visual components on large-scale image-text datasets, improving the model’s capacity to integrate and interpret multimodal information effectively. Following this evidence, the \Mississippi models employ a \textbf{pre-training} and \textbf{fine-tuning} strategy: pre-training focuses on aligning visual and textual features, while fine-tuning is dedicated to task-specific modeling. In the following sections, we describe the intent, training method and dataset distribution for the \vlmtiny and \vlm models.

\begin{figure}[h]
    \centering
    \begin{minipage}[b]{0.4\linewidth}
        \centering
        \includegraphics[width=\linewidth]{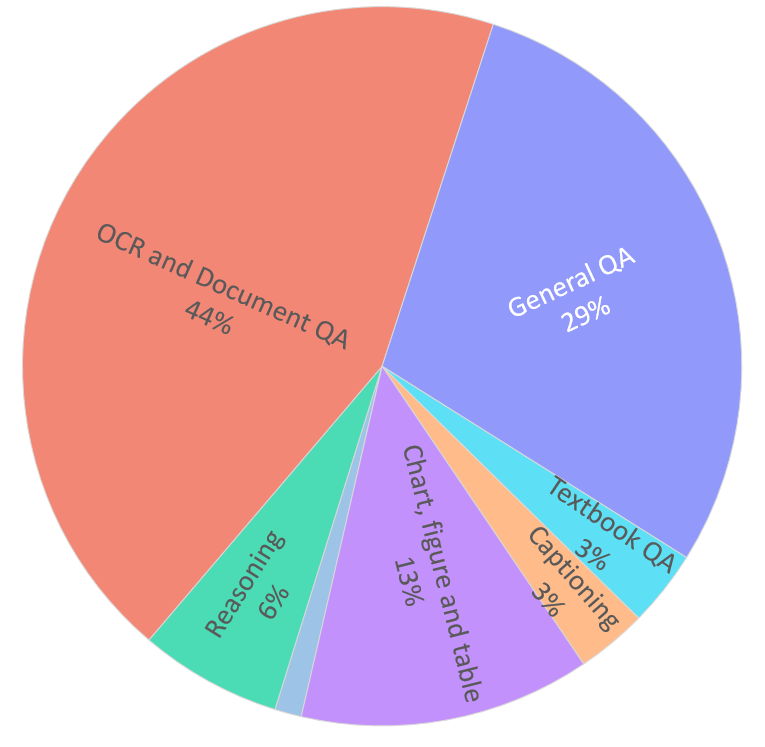}
        \subcaption{}
        \label{fig:image3}
    \end{minipage}
    \hspace{0.05\linewidth}
    \begin{minipage}[b]{0.4\linewidth}
        \centering
        \includegraphics[width=\linewidth]{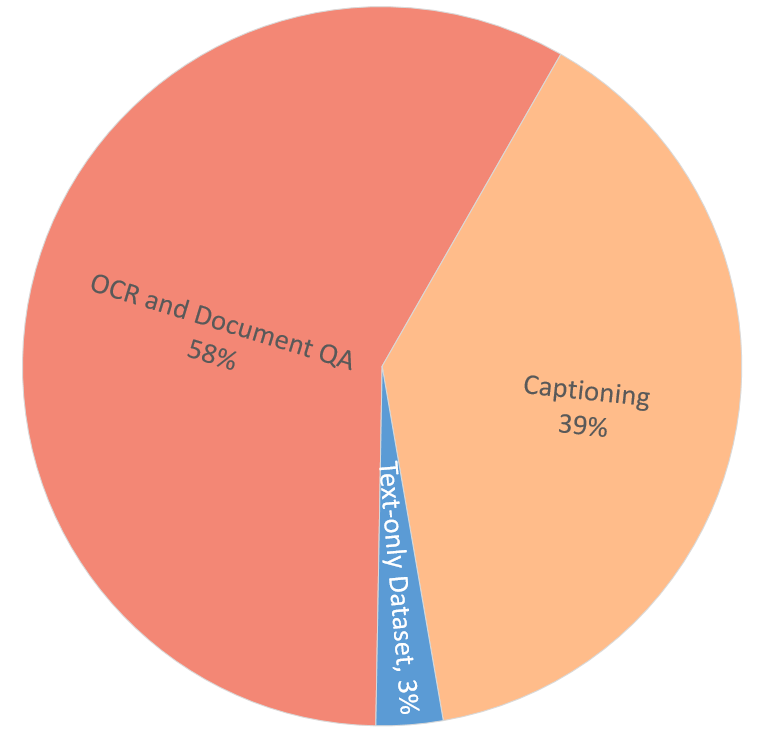}
        \subcaption{}
        \label{fig:image1}
    \end{minipage}
    \captionsetup{font=small}
    \caption{Data distribution across tasks during \textbf{pre-training} for the \Mississippi models: (a) \vlmtiny emphasizes OCR and document QA (44\%), and general QA (29\%), while (b) \vlm focuses on OCR and document QA (58\%), and captioning (39\%).}
    \label{fig:combined_pie}
\end{figure}

\begin{figure}[h]
    \centering
    \begin{minipage}[b]{0.4\linewidth}
        \centering
        \includegraphics[width=\linewidth]{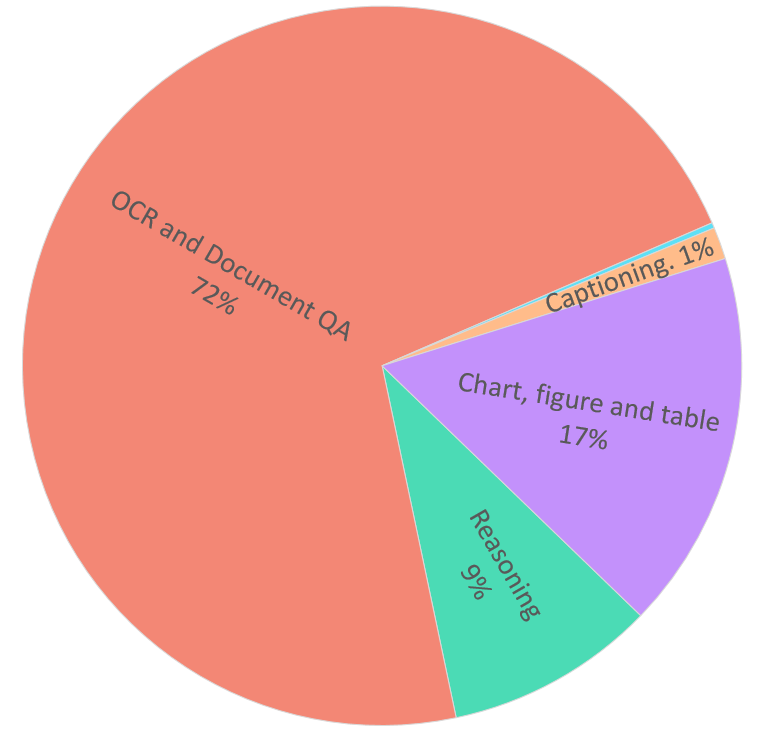}
        \subcaption{}
        \label{fig:image4}
    \end{minipage}
    \hspace{0.05\linewidth}
    \begin{minipage}[b]{0.4\linewidth}
        \centering
        \includegraphics[width=\linewidth]{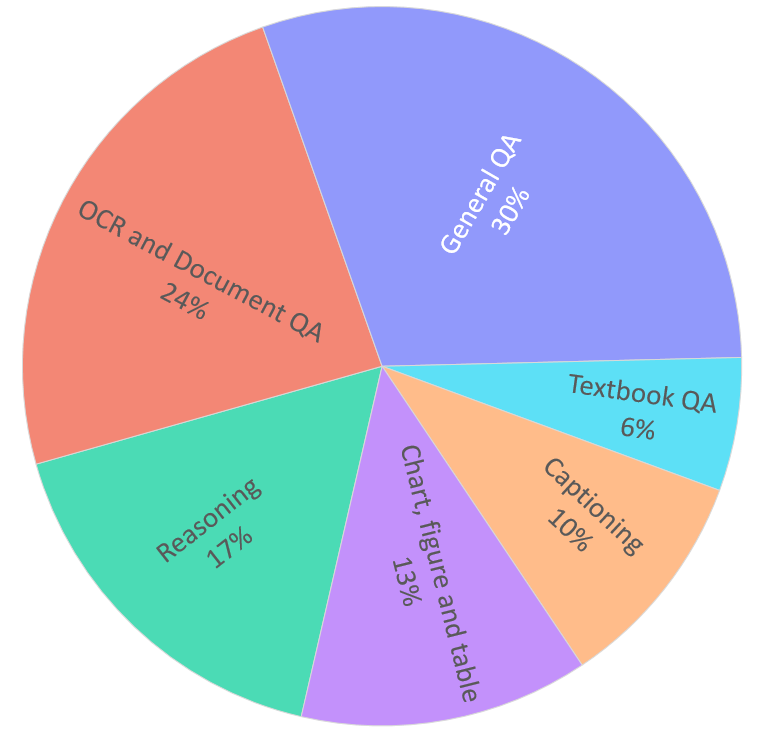}
        \subcaption{}
        \label{fig:image2}
    \end{minipage}
    \captionsetup{font=small}
    \caption{Data distribution across tasks during \textbf{fine-tuning} for the \Mississippi models: (a) \vlmtiny concentrates heavily on OCR and document QA (72\%), with chart, figure, and table tasks (17\%) as the second focus, whereas (b) \vlm balances various tasks, with general QA (39\%), reasoning (17\%), and OCR and document QA (24\%) being the key components.}
    \label{fig:combined_pie_800}
\end{figure}

\subsection{H2OVL-Mississippi-0.8B Model}
The \vlmtiny model is designed specifically for OCR and document understanding, with a focus on accurately extracting, recognizing, and interpreting text from images, particularly in complex and structured visual contexts. Its training methodology and datasets are tailored to optimize performance for these tasks.

\begin{itemize}
    \item \textbf{Pre-training}: The pre-training phase utilizes 11 million conversation examples covering a diverse range of tasks, including general QA, image captioning, OCR, and reasoning, as depicted in Figure~\ref{fig:image3}. This diverse dataset helps the model achieve a well-balanced and unbiased state, establishing a strong foundation for the subsequent OCR-specific fine-tuning. 
    The pre-training process consists of two steps. In step1, only the MLP projector is optimized, while both the ViT and LLM remain frozen, using approximately 3 percent of the pre-training dataset. In step 2, the MLP and LLM are jointly optimized, with the ViT still frozen, this time using the full pre-training dataset.
    \item \textbf{Fine-tuning}: The fine-tuning dataset consists of approximately 8 million examples, with a strong emphasis on OCR tasks such as text recognition, document parsing, and structured information extraction. To enhance the model's specialization in OCR, other general task datasets are excluded, as illustrated in Figure~\ref{fig:image4}. During this stage, all three components (ViT, MLP, and LLM) are optimized jointly.
\end{itemize}

Table~\ref{tab:data} presents the detailed data statistics, and the training hyperparameters are summarized in Table~\ref{tab:hyperparam}. For brevity, the pre-training step that focused solely on optimizing the MLP is not included.

\subsection{H2OVL-Mississippi-2B Model}
The \vlm model is designed to excel in document intelligence tasks while maintaining versatility as a general-purpose visual language model. During data composition, a significant portion (58\%) of OCR and document-related data was incorporated in pre-training to optimize document visual feature extraction and alignment. In the fine-tuning stage, we balanced the data distribution to ensure the model's performance across a diverse range of domains and tasks.

\begin{itemize}
    \item \textbf{Pre-training}: The pre-training dataset consists of 5 million conversation pairs, focusing on three key areas: OCR data, image captioning and text-only datasets. The OCR data trains the model to recognize and interpret text embedded within images, improving its skills in document understanding and text extraction from visual sources. The image captioning data connect visual inputs with corresponding textual descriptions, enhancing the model’s ability to associate images with relevant language. The text-only datasets ensure that the model maintains strong language understanding capabilities even when visual inputs are absent. The distribution of this data is illustrated in Figure~\ref{fig:image1}. During this pre-training phase, only the vision encoder and MLP projector were trained together for 4 epochs, while the LLM remained frozen.
    \item \textbf{Fine-tuning}: The fine-tuning stage of \vlm utilized 12 million conversation examples to enhance task-specific performance across various domains. The primary tasks included general question-answering (QA), which focused on handling multi-image, single-image, and text-only inputs. Additionally, OCR and document understanding were emphasized for extracting structured information from both multi- and single-image sources. Complex tasks involving reasoning, logic, and programming were also incorporated, requiring problem-solving with mixed input types. Furthermore, the fine-tuning covered captioning, textbook Q\&A, image comparison, and chart and table understanding to ensure broad task coverage and versatility, as illustrated in Figure~\ref{fig:image2}. During this stage, the full model was trained for a total of 3 epochs.
\end{itemize}

Training hyperparameters are summarized in Table~\ref{tab:hyperparam}. Data composition statistics are presented in Table~\ref{tab:data}.

\begin{table}[htbp]
\centering
\scriptsize
\captionsetup{font=small}
\caption{Summary of data for pre-training and fine-tuning of \Mississippi models}
\label{tab:data}
\begin{adjustbox}{max width=0.8\textwidth}
\begin{tabular}{llrrrrr}
\toprule
\small
\textbf{Task Composition} & \textbf{Input Type}   & \textbf{2B-pretrain} & \textbf{2B-finetune} & \textbf{0.8B-pretrain} & \textbf{0.8B-pretrain} & \textbf{0.8B-finetune} \\
& & & & \textbf{-step1}& \textbf{-step2} & \\
\midrule
General QA                                      & multi-image  &      & 332,000    &       & 332,000     &      \\
                                                & single-image &      & 1,967,797   &       & 1,781,737     &      \\
                                                & text-only    & 143,000     & 1,219,628    &      & 1,196,553     &      \\
\midrule
Reasoning, logic,            & multi-image  &      & 256,995    &       & 215,000     & 41,995     \\
maths, programming           & single-image &      & 876,245    &       & 518,929     & 705,340     \\
                                                & text-only    &      & 899,960    &       &      &      \\
\midrule
Captioning                                      & multi-image  &      & 36,000      &     & 36,000     &      \\
                                                & single-image & 1,966,936     & 1,096,585  & 196,692   & 327,756     & 113,376     \\
\midrule
OCR, document understanding, & multi-image  &      & 1,273,215    &       & 1,273,215     & 1,273,215     \\
 text transcription          & single-image & 3,141,265     & 1,626,087 & 157,063  & 3,737,088    & 4,389,070     \\
\midrule
Textbook, academic questions                    & multi-image  &      & 374,575   &        & 374,575     &      \\
                                                & single-image &      & 343,761   &        & 17,389     & 19,249     \\
\midrule
Chart, figure, table understanding              & single-image &      & 1,506,542  &         & 1,497,152     & 1,344,415     \\
\midrule
Differences between images                      & multi-image  &      & 138,000    &       & 138,000     &      \\
\bottomrule
\textbf{Total}                          & & \textbf{5,251,201} & \textbf{11,947,390} & \textbf{353,755} & \textbf{11,445,394} & \textbf{7,886,660}  
\end{tabular}
\end{adjustbox}
\end{table}

\begin{table}[htbp]
\centering
\scriptsize
\captionsetup{font=small, width=0.8\textwidth}
\caption{Hyperparameters for pre-training and fine-tuning of \Mississippi models}
\label{tab:hyperparam}
\begin{adjustbox}{max width=0.8\textwidth}
\begin{tabular}{lccccc}
\toprule
\small
\ & \textbf{2B-pretrain} & \textbf{2B-finetune} & \textbf{0.8B-pretrain-step1} & \textbf{0.8B-pretrain-step2} & \textbf{0.8B-finetune} \\
\midrule
freeze VIT  &  \ding{55}  &  \ding{55}  &  \ding{51}  &  \ding{51}  &  \ding{55}\\
freeze LLM  &  \ding{51}  &  \ding{55}  &  \ding{51}  &  \ding{55}  &  \ding{55}\\
freeze MLP  &  \ding{55}  &  \ding{55}  &  \ding{55}  &  \ding{55}  & \ding{55} \\
image size  &  448  &  448  &  448  &  448  & 448 \\
max num tiles  &  6  &  6  &  6  &  6  & 6 \\
learning rate  &  4e-5  &  4e-5 $\rightarrow$ 2e-5  &  1e-4  &  3e-5  & 1e-5 \\
scheduler  &  cosine  &  cosine  &  cosine  &  cosine  & cosine \\
batch size  &  256  &  256  &  256  &  256  & 256 \\
weight decay  &    0.01&    0.03&  0.01&  0.01  & 0.01 \\
epochs  &  4  &  2 $\rightarrow$ 1  &  1  &  1  & 1 \\
hardware  &  8 $\times$ H100  &  8 $\times$ H100  &  8 $\times$ H100  &  8 $\times$ H100  & 8 $\times$ H100 \\
hours of training  &    36&    158&  0.8&  11  & 34 \\
\bottomrule
\end{tabular}
\end{adjustbox}
\end{table}

\section{Evaluation}

In this section, we present evaluation of \Mississippi across a variety of dimensions, focusing on (1) General Vision-Language benchmarks, and (2) OCR and document-centric benchmarks.

\subsection{General Vision-Language benchmarks}

Table~\ref{tab:general_bench} provides a comprehensive comparison of models across a range of benchmarks, evaluating their strengths and weaknesses. It includes several categories of models, such as current state-of-the-art, legacy state-of-the-art, and similarly sized models. Each model’s performance is assessed using benchmarks like MMBench~\cite{MMBench}, MMStar~\cite{mmstar}, MMMU~\cite{mmmu}, Math Vista~\cite{mathvista}, Hallusion Bench~\cite{hallusionbench}, AI2D~\cite{ai2d}, OCRBench~\cite{liu2024hiddenmysteryocrlarge}, and MMVet~\cite{mmvetevaluatinglargemultimodal}, offering insights into their versatility and specialized capabilities.

Models classified under legacy state-of-the-art, such as GPT-4v (1106, detail-high) and Gemini-1.0-Pro, illustrate how quickly the field evolves. These models, though previously considered cutting-edge, now achieve lower scores, especially on advanced benchmarks like MMStar and OCRBench. For instance, GPT-4v scores 56.4 on average, with an OCRBench score of 678, which is considerably behind the newer models.

Within the category of similar size models, \vlm demonstrates competitive performance, with an average score of 54.4. \vlm excels in benchmarks like Math Vista (56.8) and OCRBench (782), positioning it as a strong model for multimodal and OCR tasks. Compared to its closest peer, Qwen2-VL-2B, \vlm shows a slight lag in benchmarks like MMBench and MMStar but remains strong in OCR-related tasks, where it outperforms several similarly sized models. The trend among similar size models highlights that while models like \vlm and Qwen2-VL may not yet reach state-of-the-art performance, they are highly effective for specific use cases such as text extraction and mathematical reasoning tasks.

We utilized VLMEvalKit\footnote{commit:e254f006fb389dc7877f64d517c14d855f7ac759} \cite{duan2024vlmevalkit} for measuring the performance of the models. For both models, we set the maximum tile number for each image to 6. Additionally, the MSAC image preprocessing function was implemented for \vlm.

\begin{table}[htbp]
\centering
\caption{Performance Comparison of Models Across Multiple Benchmarks}
\label{tab:general_bench}
\begin{adjustbox}{max width=\textwidth}
\begin{tabular}{lcccccccccc}
\toprule
\textbf{Models} & \textbf{Params (B)} & \multicolumn{9}{c}{\textbf{Benchmark Scores}} \\
\cmidrule{3-11}
 &  & \textbf{Avg. Score} & \textbf{MMBench} & \textbf{MMStar} & \textbf{MMMU\textsubscript{VAL}} & \textbf{Math Vista} & \textbf{Hallusion} & \textbf{AI2D\textsubscript{TEST}} & \textbf{OCRBench} & \textbf{MMVet} \\
 &  & \textbf{(8 Benchmarks)} & \textbf{V1.1\textsubscript{TEST}} &  &  &  & \textbf{Bench} &  &  &  \\
\midrule
\multicolumn{11}{l}{\textbf{Current state of the art}} \\
Qwen-VL-Max-0809 & 72 & \textbf{74.4}& \textbf{85.8}& \textbf{69.2}& 64.6 & \textbf{68.3}& \textbf{59.2}& \textbf{88.1}& \textbf{881}& 72.3 \\
GPT-4o-20240806 & - & 71.5 & 80.5 & 64.7 & \textbf{69.9}& 62.7 & 54.2 & 84.7 & 805 & \textbf{75.1}\\
InternVL2-Llama3-76B & 76 & 71.0 & 85.5 & 67.1 & 58.3 & 65.6 & 55.4 & 87.6 & 842 & 64.4 \\
Claude3.5-Sonnet & - & 67.9 & 78.5 & 62.2 & 65.9 & 61.6 & 49.9 & 80.2 & 788 & 66.0 \\
Gemini-1.5-Pro & - & 64.4 & 73.9 & 59.1 & 60.6 & 57.7 & 45.6 & 79.1 & 754 & 64.0 \\
\midrule
\multicolumn{11}{l}{\textbf{Legacy state of the art}} \\
GPT-4v (1106, detail-high) & - & 56.4 & 65.5 & 50.4 & 59.3 & 48.2 & 39.3 & 71.4 & 678 & 49.0 \\
Gemini-1.0-Pro & - & 56.1 & 69.7 & 38.6 & 49.0 & 46.5 & 45.7 & 72.9 & 680 & 58.6 \\
Claude3-Sonnet & - & 53.5 & 63.9 & 44.2 & 47.4 & 45.0 & 41.3 & 69.9 & 646 & 51.7 \\
Qwen-VL-Plus & - & 52.2 & 66.2 & 39.7 & 39.8 & 37.6 & 40.6 & 65.7 & 726 & 55.7 \\
\midrule
\multicolumn{11}{l}{\textbf{Similar size models}} \\
Qwen2-VL-2B & 2.1 & \textbf{57.2}& \textbf{72.2}& 47.5 & 42.2& 47.8 & \textbf{42.4}& 74.7& \textbf{797}& \textbf{51.5}\\
\textbf{H2OVL-Mississippi-2B} & 2.1 & 54.4 & 64.8 & 49.6 & 35.2 & \textbf{56.8}& 36.4 & 69.9 & 782 & 44.7 \\
InternVL2-2B & 2.1 & 53.9 & 69.6 & \textbf{49.8}& 36.3 & 46.0 & 38.0 & 74.1 & 781 & 39.7 \\
Phi-3-Vision & 4.2 & 53.6 & 65.2 & 47.7 & \textbf{46.1}& 44.6 & 39.0 & \textbf{78.4}& 637 & 44.1 \\
MiniMonkey & 2.2 & 52.7 & 68.9 & 48.1 & 35.7 & 45.3 & 30.9 & 73.7 & 794 & 39.8 \\
MiniCPM-V-2 & 2.8 & 47.9 & 65.8 & 39.1 & 38.2 & 39.8 & 36.1 & 62.9 & 605 & 41.0 \\
 InternVL2-1B& 0.8& 48.3& 59.7& 45.6& 36.7& 39.4& 34.3& 63.8& 755&31.5\\
PaliGemma-3B-mix-448 & 2.9 & 46.5 & 65.6 & 48.3 & 34.9 & 28.7 & 32.2 & 68.3 & 614 & 33.1 \\
\textbf{H2OVL-Mississippi-0.8B}& 0.8& 43.5& 47.7& 39.1& 34& 39& 29.6& 53.6& 751&30.0\\
DeepSeek-VL-1.3B & 2.0 & 39.6 & 63.8 & 39.9 & 33.8 & 29.8 & 27.6 & 51.5 & 413 & 29.2 \\
\bottomrule
\end{tabular}
\end{adjustbox}
\end{table}

\subsection{OCR and Document centric benchmarks}

\noindent\textbf{OCR Benchmarks.} We conducted a detailed comparative analysis of various vision-language models (VLMs), including the latest general OCR model (e.g., GOT-OCR2.0\cite{wei2024general}), across multiple evaluation tasks from OCRBench\cite{liu2024hiddenmysteryocrlarge}, a benchmark designed to rigorously assess OCR performance.The tasks covered include Text Recognition, Scene Text-centric VQA, Document-oriented VQA, Key Information Extraction (KIE), and Handwritten Mathematical Expression Recognition (HMER). Both \vlmtiny and \vlm demonstrated competitive performance across the board.

The \vlmtiny model stands out by achieving the highest score in OCRBench Text Recognition (274), significantly outperforming all other models, including those with much larger parameter sizes, such as InternVL2-26B and MiniCPM-V2.6. This result highlights the model’s efficiency and capability, particularly for OCR-specific tasks. Despite having fewer parameters, the 0.8B model consistently surpasses larger models in text recognition, making it an optimal choice for resource-constrained environments where high OCR performance is required.

The \vlm model also demonstrates robust performance across a range of tasks. With a total score of 782, it outperforms several models that have much larger sizes, proving its overall effectiveness. In particular, the 2B model shows competitive results in Text Recognition (252), Scene Text VQA (171), Document-Oriented VQA (140), and KIE (166), making it an excellent candidate for general document understanding and extraction tasks. 

Table~\ref{tab:ocrBench} presents detailed OCRBench results among comparable models.
To further contextualize the results, we included two traditional OCR text recognition models in our analysis: DocTR-default \cite{doctr2021} and DocTR-V2M, the latter being a retrained version developed internally by our company.

\begin{table}[htbp]
\centering
\captionsetup{font=small}
\caption{Performance Comparison of Models on OCRBench}
\label{tab:ocrBench}
\begin{adjustbox}{max width=\textwidth}
\begin{tabular}{lcllcccccc}
\toprule
 &   &  & &\multicolumn{6}{c}{\textbf{OCRBench Scores}} \\
\cmidrule{6-10}
\textbf{Models} &\textbf{Params}  & \textbf{Language Model}& \textbf{Vision Model}&\textbf{Total} & \textbf{Text} & \textbf{Scene Text} & \textbf{Document} & \textbf{KIE} & \textbf{HMER} \\
& \textbf{(B)} & & & &\textbf{Recognition} & \textbf{VQA} & \textbf{Oriented VQA} & &\\
\midrule
MiniCPM-V2.6& 8&  Qwen2-7B& SigLIP-400M&836& 259& \textbf{184}& \textbf{169}& \textbf{182}& 42\\
InternVL2-26B& 26&  internlm2-chat-20b& InternViT-6B&823& 251& \textbf{184}& 153& 168& 67\\
Qwen2-VL-2B-Instruct& 2.1&  Qwen2-1.5B& ViT-600M&812& 265& 172& 146& 174& 55\\
MiniMonkey& 2.2&  InternLM2-1.8B& InternViT-300M&792& 250& 178& 126& 171& 67\\
InternVL2-4B& 4&  Phi-3& InternViT-300M&785& 236& 170& 154& 158& 67\\
InternVL2-2B& 2&  InternLM2-1.8B& InternViT-300M&785& 246& 170& 133& 167& 69\\
\textbf{H2OVL-Mississippi-2B}& 2&  H2O-Danube2 1.8B& InternViT-300M&782& 252& 171& 140& 166& 53\\
InternVL2-1B& 0.8&  Qwen2-0.5B& InternViT-300M&755& 242& 164& 127& 150& \textbf{72}\\
\textbf{H2OVL-Mississippi-0.8B}& 0.8&  H2O-Danube3 0.5B& InternViT-300M&751& \textbf{274}& 162& 112& 152& 51\\
MiniCPM-Llama3-V2.5& 8&  Llama-3-8B-Instruct& SigLIP-400M&725& 221& 171& 125& 155& 53\\
InternVL-Chat-V1.5& 26&  InternLM2-20B& InternViT-6B&722& 236& 181& 149& 153& 3\\
Mini-InternVL-Chat-2B-V1.5& 2&  InternLM2-1.8B& InternViT-300M&652& 222& 161& 126& 139& 4\\
Phi-3-Vision& 4.2&  Phi-3& CLIP ViT-L/14&640& 196& 159& 137& 148& 0\\
Mini-InternVL-Chat-4B-V1.5& 4&  Phi-3& InternViT-300M&640& 193& 160& 146& 135& 6\\
 GOT-OCR2.0& 0.6& Qwen2-0.5B& VitDet-80M& 622& 245& 99& 83& 164&31\\
 PaliGemma-3B-mix-448& 2.9& Gemma-2B& SigLIP-400M& 613& 242& 165& 88& 118&0\\
 MiniCPM-V-2& 2.8& MiniCPM-2.4B& SigLIP-400M& 596& 243& 168& 100& 85&0\\
 DocTR-default& 0.05& -& -& -& 177& -& -& -&-\\
 DocTR-V2M(H2O.AI)& 0.05& -& -& -& 256& -& -& -&-\\
 \bottomrule
\end{tabular}
\end{adjustbox}
\end{table}

\noindent\textbf{Text Oriented VQA benchmarks.} In addition to the OCRBench evaluation, we further investigate our model’s detailed visual perception capabilities by assessing its performance on text-oriented VQA datasets, including TextVQA~\cite{singh2019vqamodelsread}, DocVQA~\cite{mathew2021docvqadatasetvqadocument}, and InfoVQA~\cite{mathew2021infographicvqa}. 
As summarized in Table \ref{tab:Public_DocAI}, \vlm demonstrates commendable overall performance across all tasks. Notably, it achieves better or comparable scores even against much larger models like Cambrian-13B (13B parameters), showing its efficiency in handling text-based VQA tasks with significantly fewer parameters. Despite its smaller size (2.1B parameters), \vlm performs competitively on TextVQA and DocVQA and demonstrates steady results on InfoVQA, underscoring the model’s robustness in diverse visual question-answering contexts.

\begin{table}[htbp]
\centering
\scriptsize  
\captionsetup{font=scriptsize}
\begin{adjustbox}{max width=\textwidth}
\begin{tabular}{lcccc}
\toprule
     \textbf{Models}&  \textbf{Params (B)}&  \textbf{TextVQA}& \textbf{DocVQA}& \textbf{InfoVQA}\\
     \midrule
     Qwen-VL-Max-0809&  72&  85.5&  96.5*& 84.5*\\
     GPT-4o-20240806& - &  70.6&  86.1& 66\\
     Qwen2-VL-2B&  2.1&  79.7&  89.2& 64.1\\
     \textbf{H2OVL-Mississippi-2B}&  2.1&  75.1&  83.8& 43\\
     InternVL2-2B&  2.1&  73.4&  86.2& 57.7\\
     MiniCPM-V-2 &  2.8&  73.2&  69.6& 38.2\\
     Cambrian-13B&  13&  72.8&  73.7& 44.3\\
     Phi-3-Vision - Microsoft&  4.2&  72.4&  84.3& 49.9\\
     PaliGemma-3B-mix-448 &  2.9&  68.1&  73.9& 34.1\\
     \bottomrule
\end{tabular}
\end{adjustbox}
\caption{\textbf{Comparison on Text-Oriented VQA.} Performance comparison with SoTA and similar sized models on public text-oriented VQA benchmarks includes: TextVQA\cite{singh2019vqamodelsread},  DocVQA\cite{mathew2021docvqadatasetvqadocument} and InfoVQA\cite{mathew2021infographicvqa}. `*` denotes numbers obtained from test set. Others are from val set.}
\label{tab:Public_DocAI}
\end{table}

\begin{table}[htbp]
\centering
\scriptsize
\captionsetup{font=scriptsize}
\begin{adjustbox}{max width=\textwidth}
\begin{tabular}{lccccc}
\toprule
 & &\multicolumn{3}{c}{\textbf{Document Type}} \\
\cmidrule(lr){3-5} 
\textbf{Models}&\textbf{Params(B)} &\textbf{Receipts}& \textbf{Drivers Licenses} & \textbf{Checks}& \textbf{Avg.}\\
\midrule
GPT-4o-20240806& -& 80.2& \textbf{81.7}& \textbf{77.6}& \textbf{79.83}\\
Claude-3-5-sonnet-20240620& -&\textbf{94.3}& \textcolor{blue}{80.5}& \textcolor{blue}{64.6}& \textcolor{blue}{79.80}\\
InternVL2-40B& 40&80.1& 68.7& 56.8& 68.53\\
InternVL2-26B& 26&76.4& 64.1& 43.4& 61.30\\
\textbf{H2OVL-Mississippi-2B}& 2.1&\textcolor{blue}{82.0}& 56.4& 41.5& 59.97\\
InternVL2-8B& 8&71.6& 60.7& 39.1& 57.13\\
MiniCPM-V-2-6& 8&62.0& 58.3& 46.3& 55.53\\
InternVL2-4B& 4&58.5& 53.4& 42.1& 51.33\\
InternVL2-2B& 2.1&60.5& 49.2& 36.5& 48.73\\
InternVL2-1B& 0.8&56.6& 41.5& 30.7& 42.93\\
\bottomrule
\end{tabular}
\end{adjustbox}
\caption{\textbf{Comparison on Information Extraction Tasks.} Performance comparison with SoTA and similarly sized models on document-specific information extraction tasks. The evaluation is conducted on real-world documents across various types, including receipts, driver’s licenses, and checks. Accuracy is reported as the average of perfect match rate, effective TED, and F1-scores regarding JSON parsing rates. The best performance for each task is highlighted in bold, while the second-best is shown in blue.}
\label{tab:DocAI_private_sets}
\end{table}

\begin{figure}[htbp] 
    \centering        
    \includegraphics[width=0.9\linewidth]{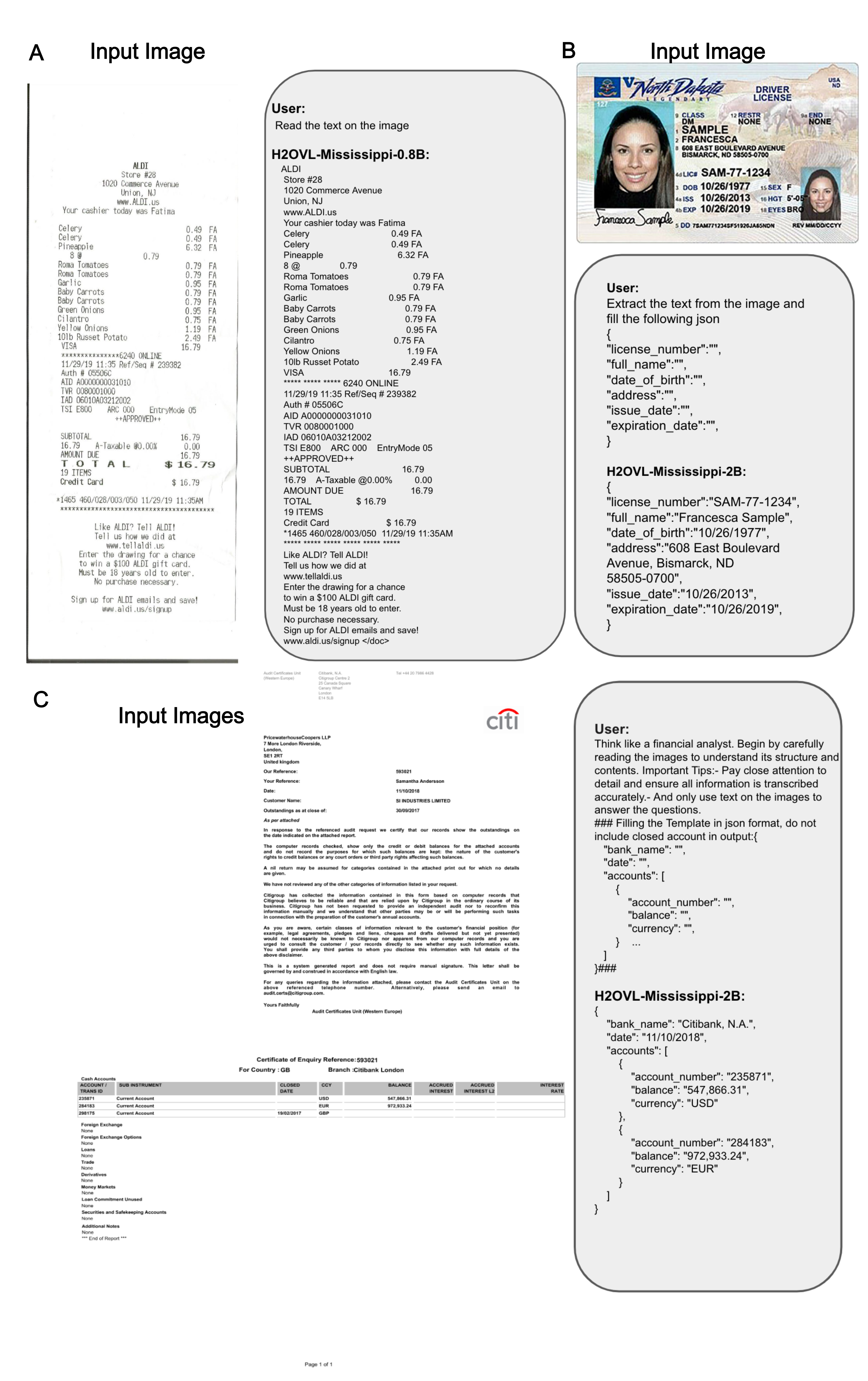} 
    \caption{Example outputs with \Mississippi models.} 
    \label{fig:example_outputs}  
\end{figure}

\noindent\textbf{Document specific information extraction benchmarks.} To further explore the document understanding capabilities of the \Mississippi models in real-world scenarios, we curated three datasets that cover some of the most common business documents: receipts, driver’s licenses, and checks. Using prompts similar to those listed in Figure~\ref{fig:example_outputs} , we evaluated the model on information extraction tasks. Performance was measured based on effective Tree Edit Distance (TED) and effective F1 derived from JSON parsing rate, and the perfect match rate. The final accuracy score was an average of these metrics for each document type.

The quantitative results are summarized in Table~\ref{tab:DocAI_private_sets}.  The \vlm model excelled in processing receipts, achieving the second-highest accuracy of 82, outperforming much larger models such as InternVL2-40B, InternVL2-26B, and GPT-4o. This underscores the model’s efficiency and strong capability in handling specific document types, despite its relatively smaller size (2.1B parameters). On driver’s licenses and checks, \vlm also showed competitive results, with scores of 56.4 and 41.5, respectively. While it did not surpass the top-performing larger models in these tasks, it outperformed some of the larger models and ranked above other similar size models.  its consistent performance across different document types highlights its effectiveness in balancing accuracy and computational efficiency.


\section{Conclusions and Future Work}
We introduce \Mississippi, a series of small language models consisting of \vlm and \vlmtiny released open source under Apache 2.0. Our models show competitive performance compared to popular models of similar size across a variety of benchmarks, including general vision-language evaluations, OCR and document-centric tasks. \Mississippi is built on our continuous efforts to contribute to the growing ecosystem of open source small language models. We are confident that our models can play a pivotal role in a wide range of applications, from typical chatting and fine-tuning for specific use cases to on-device offline applications on mobile phones or edge devices.

Through this project, we gained valuable experience in the end-to-end development of vision-language models, including data collection and preparation, input preprocessing, model architecture selection, training, and hyperparameter tuning. These learnings have prepared us to tackle more complex challenges in future work, such as:

\begin{itemize}
    \item \textbf{Improving multilingual capabilities} to extend model support for diverse languages and scripts.
    \item \textbf{Incorporating additional modalities}, such as video and audio, to enable richer multimodal understanding.
    \item \textbf{Scaling up model sizes} to 4B, 7B, or even larger, to further enhance performance and address more complex tasks.
    \item \textbf{Addressing agent-based tasks} that involve decision-making and real-world interaction, enabling the models to function effectively in dynamic environments.
    \item \textbf{Enhancing fine-grained visual capabilities} to improve performance in tasks that require distinguishing between highly similar objects or parsing intricate scenes.
\end{itemize}

\clearpage

\clearpage
\small

\section*{Acknowledgments} 
We would like to thank Philipp Singer and Yauhen Babakhin of H2O.AI Danube team for data contribution and insightful discussion.

\bibliographystyle{unsrt}
\bibliography{references}   

\end{document}